\documentclass[11pt,letterpaper]{article}
\usepackage{amssymb,amsmath}
\usepackage{hyperref}
\usepackage{natbib}
\usepackage{times}
\usepackage{xcolor}

\title{Faster Transformer Decoding: N-gram Masked Self-Attention}
\author{Ciprian Chelba and Mia Chen and Ankur Bapna and Noam Shazeer\\
  Google, Inc.\\
  1600 Amphitheatre Parkway\\
  Mountain View, CA 94043, USA\\
  {\tt \{ciprianchelba,miachen,ankurbpn,noam\}@google.com}}

\date{December 19, 2019}

\begin{document}
\maketitle

\begin{abstract}
    Motivated by the fact that most of the information relevant to the prediction of target tokens is drawn from the source sentence $S=s_1, \ldots, s_S$, we propose truncating the target-side window used for computing self-attention by making an $N$-gram assumption. Experiments on WMT EnDe and EnFr data sets show that the $N$-gram masked self-attention model loses very little in BLEU score for $N$ values in the range $4, \ldots, 8$, depending on the task.
\end{abstract}

\section{Introduction}
Transformers~\citep{vaswani2017attention} are the most effective neural architectures for sequence modeling problems encountered in natural language, in particular language modeling and machine translation (MT). For MT in particular, the most successful modeling paradigm predicts a given target word using a conditional probability model leveraging all source words and the previous target words. 

A first empirical observation is that the perplexity of a language model (LM) that predicts the target sentence using $P(t_k|t_1, \ldots, t_{k-1}; \theta_{LM})$ is significantly higher than that of a conditional neural translation model (NMT): $P(t_k|t_1, \ldots, t_{k-1},  s_1, \ldots, s_{S}; \theta_{NMT})$. A transformer NMT model (6 layers, 8 attention heads, 512 model/embedding, 2048 hidden dimensionality, respectively, dropout 0.1) trained on the WMT EnDe data used for quality evaluation as described in Section 4.1.1 of~\citep{wang2018qe} achieves conditional perplexity (PPL) 13.5 on the newstest2017 test data. An LSTM LM (2-layer, 1024 embedding and 2048 state dimensionality, respectively) trained on the De monolingual side of the parallel data, using the same word-piece model/vocabulary as the NMT model (32k, bilingual), achieves PPL 99.5.

Motivated by the fact that most of the information relevant to the prediction of target token $t_k$ is drawn from the source sentence $S=s_1, \ldots, s_S$, we propose truncating the target-side window used for computing self-attention by making an $N$-gram assumption. The self-attention mechanism in transformer models~\citep{vaswani2017attention} already employs masking to make sure only target tokens prior to the current predicted position $k$ are used when estimating $P(t_k|t_1, \ldots, t_{k-1}; s_1, \ldots, s_{S}; \theta_{NMT})$. Our proposed $N$-gram mask will restrict the self-attention mechanism to using only the previous $N-1$ tokens.

A more detailed description of the $N$-gram self-attention mechanism is presented in Section~\ref{sec:ngram_sa}. The experiments presented in Section~\ref{sec:exps} compare the baseline with the $N$-gram self-attention transformer.

\section{N-gram Self-attention} \label{sec:ngram_sa}

The incremental computation of encodings for the target context $t_1, \ldots, t_{k-1}$ in a layered transformer decoder involves the following steps at each layer, after receiving the token embeddings (along with position encoding) or the encodings from the previous layer:
\begin{enumerate}
    \item compute the self-attention query $q_k$ and key $k_k$
    \item compute softmax over the indexes $j = 1, \ldots, k-1$, and then the attention context for position $k$, of $\mathcal{O}(k)$
    \item feed-forward computation of encoding for position 
    $k$
\end{enumerate}

For a target sentence of length $T$, the above steps need to be repeated $\forall k = 1, \ldots, T$, resulting in computational complexity of $\mathcal{O}(T^2)$.

The $N$-gram self-attention mechanism reduces the context used for the prediction at position $k$, resulting in a computational complexity of $\mathcal{O}(N \cdot T)$. As shown in our experiments, see Section~\ref{sec:exps}, $N=8$ is a viable value for the $N$-gram order; for sentences of length $T \approx 16-25$ this promises a reduction in computational complexity on the order of $\mathcal{O}(T/N)$, or an $\approx 2-3X$ speed-up for the self-attention computation. 

It remains to be seen to what extent this can be realized in practice in a given implementation and hardware platform (CPU, GPU or TPU), since the sibling feed-forward computation at step 3 may dominate the incremental computation at position $k$ in the target sentence. The results in~\citep{zhang2018accelerating} do show that optimizing the self-attention computation can have a significant impact on decoding speed.

Another potential computational advantage is the ability to store the context in a fixed size memory buffer of length $N-1$, replacing context elements one by one as the decoder advances in the target sentence, e.g. by indexing the buffer modulo $N-1$. This  reduces the memory bandwidth required by the model at inference/beam-search time (a major bottleneck on TPU) by a factor of $\mathcal{O}(T/N)$.


\section{Experiments} \label{sec:exps}

We implemented $N$-gram self-attention in lingvo~\citep{shen2019lingvo} as a configuration option to \href{https://github.com/tensorflow/lingvo/blob/cab4bce76bd485399f30a96df04ab69146859429/lingvo/core/layers_with_attention.py#L65}{TransformerAttentionLayer} (\verb+lingvo/core/layers_with_attention.py+). We perform experiments on two data sets: WMT'18 EnDe and WMT'14 EnFr; for EnDe we use newstest2012/2017 as dev/test data, respectively; for EnFr we concatenate newstest2012 and newstest2013 as dev data and use newstest2014 as test data. 

The EnDe transformer model used is configured as follows: 6 layers, 8 attention heads, 512 model/embedding, 2048 hidden dimensionality, respectively, dropout 0.1. For EnFr we used 6 layers, 16 attention heads, 1024 model/embedding, 8192 hidden dimensionality, respectively, dropout 0.1. In both cases we used bilingual (source, target) word-piece models of size 32k.

The results are presented in Tables~\ref{tab:en_de}-\ref{tab:en_fr}. Setting $N = 8$ or $N = 10$ achieves the best BLEU score on dev data and is within 0.3-0.4 BLEU from the baseline. Smaller $N$-gram orders are also a viable choice since performance degrades gracefully for $N \geq 3$.



\begin{table}[ht]
    \centering
    \begin{tabular}{|l|r|r|r|r|l|}\hline
         Model &  \multicolumn{2}{l|}{corpus BLEU} &  \multicolumn{2}{l|}{log\_pplx} & @steps \\
               & dev & test & dev & test & (dev)\\\hline
    baseline & 22.4 & 28.6 & 2.96 & 2.64 & @151.6k\\
    3-gram  &  22.1 & 28.0 & 3.12 & 2.76 & @98.32k\\
    4-gram  &  22.2 & 28.4 & 3.06 & 2.72 & @93.62k\\
    6-gram  &  22.2 & 28.3 & 3.01 & 2.68 & @347.5k\\
    8-gram  &  22.5 & 28.2 & 2.99 & 2.66 & @467.8k\\
    10-gram  &  22.5 & 28.3 & 2.98 & 2.65 & @126.9k\\\hline
    \end{tabular}
    \caption{WMT'18 EnDe experiments: corpus BLEU and log-perplexity on the dev/test data, respectively at the best checkpoint picked according to the BLEU score on dev data.}
    \label{tab:en_de}
\end{table}

%
\begin{table}[ht]
    \centering
    \begin{tabular}{|l|r|r|r|r|l|}\hline
         Model &  \multicolumn{2}{l|}{corpus BLEU} &  \multicolumn{2}{l|}{log\_pplx} & @steps \\
               & dev & test & dev & test & (dev)\\\hline
    baseline & 32.9 & 41.1/40.7+ & 2.52 & 2.16 & @175.4k\\
    2-gram   & 30.0 & 36.7       & 2.97 & 2.52 & @176.4k\\
    3-gram   & 32.2 & 40.3       & 2.58 & 2.21 & @183.7k\\
    4-gram   & 32.4 & 40.4       & 2.56 & 2.19 & @180.3k\\
    6-gram   & 32.7 & 41.1       & 2.53 & 2.17 & @200.4k\\
    8-gram   & 32.7 & 40.7       & 2.52 & 2.17 & @176.5k\\
    10-gram  & 32.8 & 40.8       & 2.52 & 2.17 & @166.9k\\
\hline
    \end{tabular}
    \caption{WMT'14 EnFr experiments: corpus BLEU and log-perplexity on the dev/test data, respectively at the best checkpoint picked according to the BLEU score on dev data; +: depending on the exact checkpoint on test data.}
    \label{tab:en_fr}
\end{table}

\section{Related Work} \label{sec:related_work}

The use of local attention mechanism is not novel. It is used in LM and NMT experiments reported in~\citep{shazeer2019fast}, text summarization work in~\citep{liu2018wikipedia}, image processing in~\citep{parmar2018image} and as well as automatic speech recognition as in~\citep{povey}. We wish to clarify that our use of local attention is restricted to the decoder component of the transformer model, unlike the image/speech processing use cases highlighted previously. 

Unlike the blocking algorithm described in Section 4.2.4 of~\citep{liu2018wikipedia}, we incrementally slide the attention window, just as also implemented in~\citep{shazeer2019fast}, see footnote on page 7 and results for $N$ = 32 in Tables 1 and 2, rows labeled with ``-local'' suffix. Our experiments show that significantly smaller values for the window length $N$ are feasible.

A detailed performance analysis in both number of operations and memory footprint is presented in~\citep{shazeer2019fast}; particularly relevant is the one in Section 3.1 for incremental operation at inference time. Directly relevant to our proposed self-attention variant is the one in \citep{zhang2018accelerating} showing 4X improvements in decoding speed over the baseline transformer model. 

\section{Conclusion and Future Work}

Experimental results show that $N$-gram self-attention is a viable alternative to full (one-sided/causal) self-attention in the decoder component of transformer models used in NMT. This slight model change promises run-time advantages in terms of memory footprint and speed that are yet to be investigated thoroughly, particularly at inference time.

\bibliography{faster_transformer_decoding}

\begin{thebibliography}{8}
\expandafter\ifx\csname natexlab\endcsname\relax\def\natexlab#1{#1}\fi

\bibitem[{Liu et~al.(2018)Liu, Saleh, Pot, Goodrich, Sepassi, Kaiser, and
  Shazeer}]{liu2018wikipedia}
Peter~J. Liu, Mohammad Saleh, Etienne Pot, Ben Goodrich, Ryan Sepassi, Lukasz
  Kaiser, and Noam Shazeer. 2018.
\newblock {Generating Wikipedia by Summarizing Long Sequences}.
\newblock \emph{CoRR}, abs/1801.10198.

\bibitem[{Parmar et~al.(2018)Parmar, Vaswani, Uszkoreit, Łukasz Kaiser,
  Shazeer, Ku, and Tran}]{parmar2018image}
Niki Parmar, Ashish Vaswani, Jakob Uszkoreit, Łukasz Kaiser, Noam Shazeer,
  Alexander Ku, and Dustin Tran. 2018.
\newblock {Image Transformer}.
\newblock In \emph{Proceedings of the 35th International Conference on Machine
  Learning}.

\bibitem[{Povey et~al.(2018)Povey, Hadian, Ghahremani, Li, and
  Khudanpur}]{povey}
Daniel Povey, Hossein Hadian, Pegah Ghahremani, Ke~Li, and Sanjeev Khudanpur.
  2018.
\newblock {A Time-restricted Self Attention Layer for ASR}.
\newblock In \emph{Proceedings of the International Conference on Acoustics,
  Speech and Signal Processing (ICASSP)}. IEEE.

\bibitem[{Shazeer(2019)}]{shazeer2019fast}
Noam Shazeer. 2019.
\newblock {Fast Transformer Decoding: One Write-Head is All You Need}.
\newblock \emph{ArXiv}.

\bibitem[{Shen et~al.(2019)Shen, Nguyen, Wu, Chen, Chen, Jia, Kannan, Sainath,
  and et~al.}]{shen2019lingvo}
Jonathan Shen, Patrick Nguyen, Yonghui Wu, Zhifeng Chen, Mia~X. Chen, Ye~Jia,
  Anjuli Kannan, Tara~N. Sainath, and Yuan~Cao et~al. 2019.
\newblock {Lingvo: a Modular and Scalable Framework for Sequence-to-Sequence
  Modeling}.
\newblock \emph{CoRR}, abs/1902.08295.

\bibitem[{Vaswani et~al.(2017)Vaswani, Shazeer, Parmar, Uszkoreit, Jones,
  Gomez, Kaiser, and Polosukhin}]{vaswani2017attention}
Ashish Vaswani, Noam Shazeer, Niki Parmar, Jakob Uszkoreit, Llion Jones,
  Aidan~N Gomez, {\L}ukasz Kaiser, and Illia Polosukhin. 2017.
\newblock \href
  {https://papers.nips.cc/paper/7181-attention-is-all-you-need.pdf} {{Attention
  Is All You Need}}.
\newblock In \emph{{Advances in Neural Information Processing Systems}}, pages
  5998--6008.

\bibitem[{Wang et~al.(2018)Wang, Fan, Li, Zhou, Chen, Shi, and Si}]{wang2018qe}
Jiayi Wang, Kai Fan, Bo~Li, Fengming Zhou, Boxing Chen, Yangbin Shi, and Luo
  Si. 2018.
\newblock \href {http://www.aclweb.org/anthology/W18-6465} {{Alibaba Submission
  for {WMT18} Quality Estimation Task}}.
\newblock In \emph{Proceedings of the Third Conference on Machine Translation:
  Shared Task Papers}, pages 809--815, Belgium, Brussels.

\bibitem[{Zhang et~al.(2018)Zhang, Xiong, and Su}]{zhang2018accelerating}
Biao Zhang, Deyi Xiong, and Jinsong Su. 2018.
\newblock {Accelerating Neural Transformer via an Average Attention Network}.
\newblock \emph{CoRR}, abs/1805.00631.

\end{thebibliography}
\bibliographystyle{acl_natbib}
\end{document}